\documentclass{article}
\usepackage{multicol}
\usepackage{PRIMEarxiv}

\usepackage[utf8]{inputenc}   % allow utf-8 input
\usepackage[T1]{fontenc}      % use 8-bit T1 fonts
\usepackage{url}              % simple URL typesetting
\usepackage{booktabs}         % professional-quality tables
\usepackage{amsfonts}         % blackboard math symbols
\usepackage{nicefrac}         % compact symbols for 1/2, etc.
\usepackage{fancyhdr}         % header
\usepackage{graphicx}         % graphics
\usepackage{amsmath}
\usepackage{inconsolata}
\usepackage{microtype}        % microtypography
\usepackage{hyperref}         % hyperlinks (load last to avoid conflicts)

\providecommand{\Description}[1]{}

\title{RH-RAG: Trustworthy Long-Form Generation for Privacy-Constrained Settings}

\author{
  Raj Shekhar Singh \\
  Indian Institute of Technology, Roorkee \\
  Roorkee, Uttarakhand, India \\
  \texttt{raj\_ss@ece.iitr.ac.in}
}

\begin{document}
\maketitle

\begin{abstract}
Generating long-form content from confidential internal reports is challenging for organizations under strict privacy constraints, where proprietary cloud APIs are infeasible. Locally deployed open-weight models offer a privacy-preserving alternative but suffer from hallucinations, weak global planning, and semantic drift over extended outputs. We present \textbf{RH-RAG}, a multi-agent framework for secure and trustworthy long-form generation using local language models. RH-RAG decomposes generation into three coordinated stages: a \emph{Planner Agent} that constructs a global outline from semantic summaries, a \emph{Writer Agent} that generates coherent section-wise content via bounded coherence memory, and a \emph{Checker Agent} that mitigates hallucinations through NLI-based verification and an attestation-driven revision loop. A dual-level retrieval index supports efficient planning and fine-grained contextual generation on consumer-grade hardware. Evaluations across literary, financial, and legal domains show that RH-RAG consistently improves factual grounding, semantic coherence, and document-level alignment over RAG baselines, approaching the reliability of proprietary cloud systems without compromising data privacy.
\end{abstract}
\begin{multicols}{2}
%%
%% SECTION 1: Introduction
%%
\section{Introduction}
Large Language Models have transformed content generation and reasoning, yet cloud-hosted proprietary APIs are infeasible for organizations under strict privacy requirements. Healthcare, finance, and legal sectors produce confidential reports that cannot be sent to third-party providers, constraining them to locally deployed 7-8B open-weight models. Long-form generation with these models compounds three failure modes: \textit{factual hallucination} (unsupported claims accumulate over extended outputs), \textit{context saturation} (retrieval fidelity degrades as context grows), and \textit{structural drift} (absence of global planning erodes cross-section coherence). Retrieval-Augmented Generation (RAG)~\cite{lewis2020rag,izacard2021leveraging} partially mitigates hallucination but lacks global planning and degrades narrative continuity over extended documents~\cite{edge2024graphrag}.

These challenges are further compounded by how smaller models utilize context. Even with expanded context windows, local models tend to attend to superficial statistical patterns rather than accurately integrating relevant retrieved passages. Conventional RAG pipelines retrieve a flat pool of evidence and generate the entire response in a single pass, causing the model to drift off-topic, repeat content, or introduce contradictory claims in later sections. Multi-step generation strategies have been proposed to address structural coherence~\cite{fierro2024planningcitations,shao2024storm}, but they are not designed for local deployment and provide no mechanism for deterministic factual verification after each generated segment.

We propose \textbf{RH-RAG}, a multi-agent framework that replicates the human writing workflow---plan, draft, verify---using locally deployed open-weight models. Documents are organized into a \textit{dual-level index} of routing summaries and evidence chunks; a \textit{Planner Agent} constructs a global outline; a \textit{Writer Agent} generates section-wise content with bounded coherence memory; and a \textit{Checker Agent} enforces factual grounding through an NLI-based attestation loop, entirely within private local environments. Contributions: \textbf{(1)} a dual-level retrieval architecture for efficient multi-agent retrieval on constrained hardware; \textbf{(2)} section-wise generation with local+global coherence memory to prevent context saturation; \textbf{(3)} an attestation-driven revision loop using a coverage-based factuality metric to systematically eliminate unsupported claims.

%%
%% SECTION 2: Related Work
%%
\begin{figure*}[t]
    \centering
    \includegraphics[width=\textwidth]{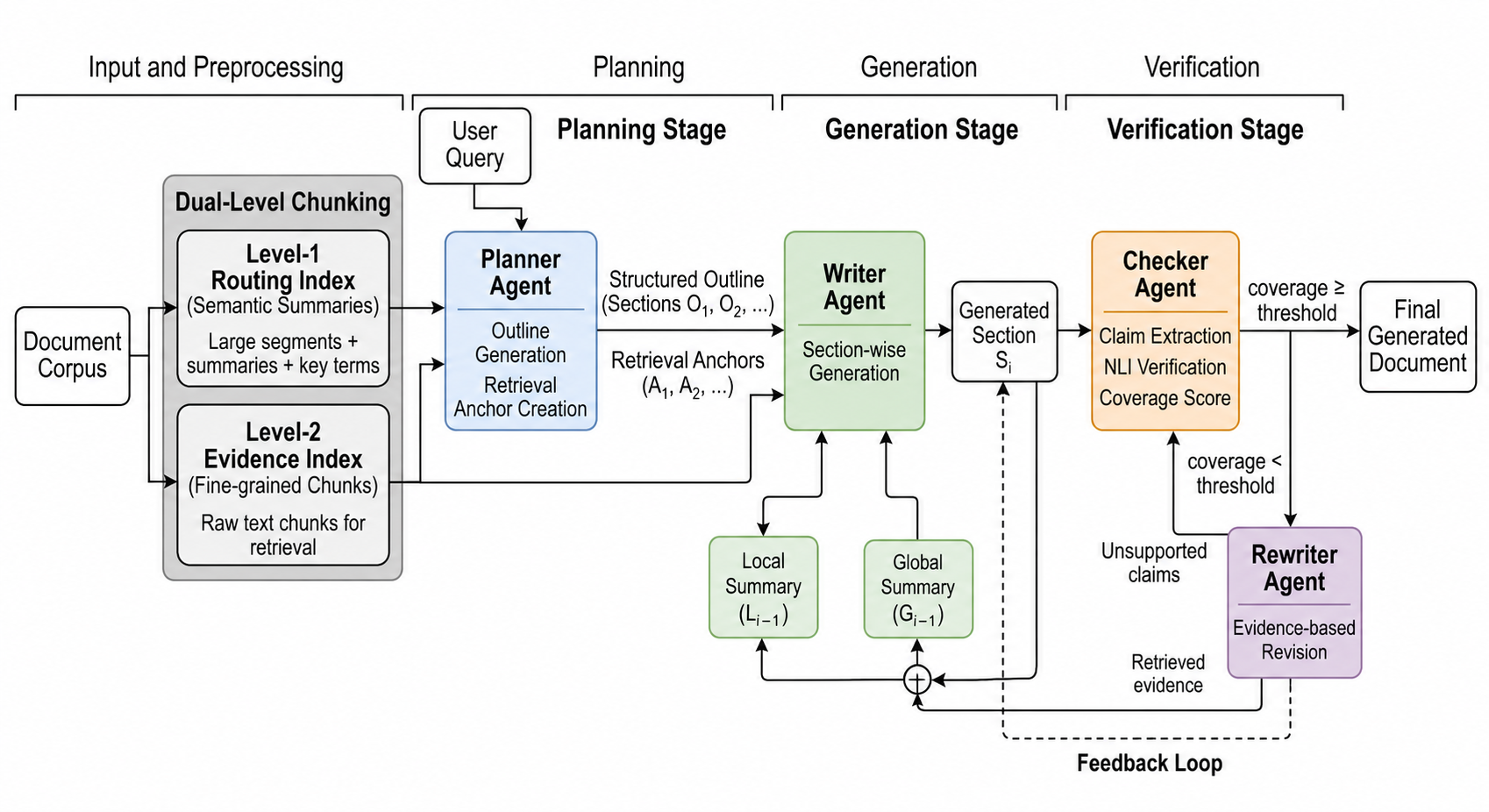}
    \caption{RH-RAG framework: dual-level chunking, role-conditioned section-wise generation with bounded coherence memory, and the NLI-backed attestation feedback loop.}
    \Description{Overview diagram of the RH-RAG multi-agent pipeline.}
    \label{fig:architecture}
\end{figure*}

\section{Related Work}

\paragraph{RAG and Structured Retrieval.}
Standard RAG pipelines retrieve passages via dense embeddings for single-pass generation~\cite{lewis2020rag,izacard2021leveraging,shi2023replug}. GraphRAG~\cite{edge2024graphrag} and hierarchical RAG~\cite{song2024hierarchicalcontext} add graph-structured and multi-granularity retrieval respectively, while MindMap~\cite{wen2024mindmap} and MIRAGE~\cite{mirage2026scaling} incorporate multi-hop graph traversals for complex queries. These approaches optimize factual precision over structured knowledge bases but do not address global narrative coherence in free-form long-form generation. Multi-step generation strategies~\cite{fierro2024planningcitations,asai2023selfrag,shao2024storm} improve structure by decomposing into planning and drafting stages, yet none integrate deterministic claim-level NLI verification within a privacy-preserving local deployment setting.

\paragraph{Graph-Based and Multi-Hop Retrieval.}
A more recent line of work moves beyond flat chunk retrieval by organizing knowledge into explicit graph structures. GraphRAG~\cite{edge2024graphrag} constructs entity-relation graphs from source documents and uses community detection to produce hierarchical summaries, enabling query-focused summarization that captures global document themes rather than locally similar passages. MindMap~\cite{wen2024mindmap} integrates structured knowledge graphs directly into the prompting process, unifying the model's parametric knowledge with explicit relational evidence to trigger a graph-of-thoughts reasoning chain. MIRAGE~\cite{mirage2026scaling} advances this further by establishing a test-time scalable inference framework that decomposes high-level prompts into entity-grounded sub-questions, executes parallel multi-hop graph traversals, and applies cross-chain verification to resolve logical contradictions across retrieved reasoning paths. While these approaches substantially improve multi-step factual precision and traceability over well-structured ontological databases, they face a critical limitation for our setting: adapting graph-based workflows to unstructured raw document text requires pre-built knowledge graphs that are expensive to construct and maintain, and their retrieval mechanisms are optimized for answering discrete factual queries rather than sustaining global narrative coherence across a multi-section long-form document.

\paragraph{Factual Verification as Generative Control.}
FActScore~\cite{min2023factscore}, AlignScore~\cite{zha2023alignscore}, and SUMMAC~\cite{laban2021summac} decompose outputs into atomic claims and verify them against source documents for offline evaluation. RH-RAG internalizes this principle as a \textit{generative control mechanism}: the Checker Agent applies NLI-based verification not merely to measure quality post-hoc but to actively trigger targeted revisions during generation, directly closing the loop between factuality measurement and output improvement.

Taken together, existing work leaves an important gap: no prior system simultaneously addresses global document planning, bounded-memory section-wise generation, deterministic claim-level verification, and strict local-only deployment within a single unified framework. RH-RAG is designed specifically to fill this gap.

%%
%% SECTION 3: RH-RAG Framework
%%
\section{The RH-RAG Framework}

Figure~\ref{fig:architecture} provides a full overview of the RH-RAG pipeline. The framework operates in three sequential stages---preprocessing, generation, and verification---each handled by a dedicated agent. Full prompt templates for all agents are provided in Appendix~\ref{app:prompts}.

\subsection{Dual-Level Document Chunking}

To balance global context understanding with precise factual retrieval, documents are indexed at two complementary granularities.

\textbf{Level-1 (Routing Summaries):} Each document is divided into large segments of 20{,}000--30{,}000 tokens, chosen to provide a good trade-off between semantic coherence and retrieval granularity. A concise summary is generated per segment using a Mistral-7B-Instruct model fine-tuned on XSum, augmented with extracted key terminologies. These summaries form a lightweight semantic routing layer used exclusively during the planning stage to identify relevant document regions without loading the full corpus into context.

\textbf{Level-2 (Evidence Chunks):} Fine-grained passages preserve the original document text for factual grounding. These chunks are encoded using the \texttt{bge-base-en-v1.5} embedding model to produce dense semantic representations that support efficient similarity-based retrieval during both generation and verification. Structural metadata (section provenance) is preserved alongside each chunk to maintain document organization during retrieval.

This separation of planning-level and generation-level indices is the architectural foundation that allows each agent to operate with precisely the granularity of context it requires, avoiding the context saturation that afflicts flat single-index RAG pipelines.

\subsection{Planner Agent}

The Planner Agent is responsible for constructing a structured document outline before any text is produced, analogous to the pre-writing planning stage in professional document composition. It queries the Level-1 routing index using the user query to retrieve high-level semantic summaries, identifies key themes and their relative importance, and organizes them into a logically ordered sequence of sections forming the global document structure.

Critically, for each section the planner produces a \textit{retrieval anchor}---a targeted query specifically designed to retrieve the most relevant Level-2 evidence chunks during the subsequent writing stage. These anchors decouple global structural planning from fine-grained evidence retrieval: the planner reasons at the level of document organization, while the writer reasons at the level of local evidence grounding. This separation ensures that narrative coherence is established before generation begins, rather than emerging (or failing to emerge) as an afterthought.

\subsection{Writer Agent}

The Writer Agent generates document content incrementally, producing one section at a time according to the planned outline. For section $i$, the agent receives the section outline $O_i$, the top-$k$ Level-2 evidence chunks $R_i$ retrieved via the planner's anchor $A_i$, and a compact bounded coherence memory $M_{i-1}$:
\begin{equation}
S_i = \text{Writer}(O_i, R_i, L_{i-1}, G_{i-1})
\end{equation}

The \textbf{bounded coherence memory} is a dual-component structure that avoids the context saturation caused by accumulating all previously generated text. It consists of: (i) a \textit{local summary} $L_{i-1}$---one to two sentences capturing the immediate narrative transition from the previous section, and (ii) a \textit{global summary} $G_{i-1}$---a running compressed representation of the entire document generated so far, preserving overarching thematic arc and key established facts. Both components are maintained by a lightweight Mistral-7B-Instruct summarizer fine-tuned on XSum. After each section $S_i$ is accepted, $L_i$ is generated directly from $S_i$, and $G_i$ is produced by updating $G_{i-1}$ with the new content. This dual-track update ensures that both local continuity (preventing abrupt narrative transitions between adjacent sections) and global coherence (preventing thematic drift across the full document) are preserved without growing the effective context size.

\subsection{Checker Agent and Attestation Feedback Loop}

Each generated section $S_i$ undergoes factual verification before incorporation into the final document. The verification pipeline operates in three stages.

\textbf{Claim extraction.} The checker first applies a rhetorical filtering step to remove transitional and stylistic sentences, then uses a Mistral-7B-Instruct model to decompose the remaining text into atomic factual claims $C_i = \{c_1, \ldots, c_m\}$, resolving co-references and splitting compound assertions to ensure each claim is independently verifiable.

\textbf{NLI verification.} For each claim $c_j \in C_i$, the top-$k$ Level-2 evidence passages are retrieved via dense similarity search, and a DeBERTa-v3-large NLI model classifies the claim as \textit{supported}, \textit{unsupported}, or \textit{contradicted} given the retrieved evidence. The section's factual coverage is then quantified as:
\begin{equation}
\text{coverage}(S_i) = \frac{|C_{\text{supported}}|}{|C_{\text{total}}|}
\end{equation}

\textbf{Attestation feedback loop.} If $\text{coverage}(S_i) \geq \tau$, the section is accepted. Otherwise, unsupported and contradicted claims are rewritten as targeted retrieval queries to fetch additional evidence, and a dedicated \textbf{Revision Agent} produces a corrected section:
\begin{equation}
S_i^{(t+1)} = \text{RevisionAgent}(S_i^{(t)},\; U_i,\; E^{\text{new}})
\end{equation}
where $U_i$ is the set of failed claims and $E^{\text{new}}$ is the newly retrieved evidence. Unlike the Writer Agent, the Revision Agent targets only the unsupported sentences, preserving surrounding paragraph structure. This loop runs for up to $T{=}3$ iterations with $k{=}5$ passages per claim; we use $\tau{=}0.8$ as the default threshold based on the ablation in Table~\ref{tab:threshold} (detailed analysis in Appendix~\ref{app:ablation}).

Table~\ref{tab:main} presents results across all domains and models.

\begin{table*}[t]
\centering
\caption{Performance across literary, financial, and legal domains. WR: win rate; Align: AlignScore.}
\label{tab:main}
\small
\begin{tabular}{lccc|ccc|ccc}
\toprule
& \multicolumn{3}{c|}{Literary} & \multicolumn{3}{c|}{Financial} & \multicolumn{3}{c}{Legal} \\
Model & WR & SUMMAC & Align & WR & SUMMAC & Align & WR & SUMMAC & Align \\
\midrule
Standard RAG           & 4.3 & 40.9 & 72.9 & 4.6 & 41.4 & 72.4 & 4.5 & 40.2 & 71.2 \\
Hierarchical RAG       & 5.7 & 46.3 & 71.3 & 5.4 & 43.2 & 73.1 & 5.6 & 44.7 & 72.6 \\
ChatGPT 4o             & 7.8 & 49.1 & 79.1 & 7.8 & 48.7 & 78.8 & 7.9 & 48.3 & 77.6 \\
Claude                 & 8.1 & 48.3 & 78.7 & 8.2 & 48.9 & 78.7 & 8.4 & 48.6 & 77.2 \\
Gemini Pro             & 7.9 & 48.8 & 79.3 & 7.8 & 48.7 & 79.1 & 7.9 & 48.4 & 77.8 \\
RH-RAG (w/o checker)  & 6.3 & 41.7 & 73.3 & 6.4 & 42.8 & 74.2 & 6.5 & 41.4 & 73.7 \\
RH-RAG (Mistral-7B)   & 6.8 & 47.6 & 76.7 & 6.7 & 48.7 & 78.5 & 6.9 & 47.5 & 78.3 \\
RH-RAG (LLaMA-3.1-8B) & 7.1 & 46.4 & 76.7 & 6.9 & 48.7 & 78.7 & 7.0 & 47.8 & 78.6 \\
RH-RAG (Qwen-2.5-7B)  & 7.2 & 47.4 & 76.7 & 7.0 & 49.4 & 78.5 & 6.8 & 47.8 & 78.4 \\
\bottomrule
\end{tabular}
\end{table*}
\begin{table}[t]
    \centering
    \caption{Impact of coverage threshold $\tau$ on factual alignment (RH-RAG Mistral-7B).}
    \label{tab:threshold}
    \begin{tabular}{lcc}
    \toprule
    Threshold & SUMMAC & AlignScore \\
    \midrule
    $\tau = 0.7$ & 46.5 & 77.6 \\
    $\tau = 0.8$ & 47.2 & 77.9 \\
    $\tau = 0.9$ & 47.9 & 78.1 \\
    \bottomrule
    \end{tabular}
\end{table}
%%
%% SECTION 4: Experiments
%%
\section{Experiments}

\subsection{Dataset}

We construct a custom multi-domain evaluation corpus (full statistics in Appendix~\ref{app:dataset}) spanning three document categories designed to stress-test different aspects of long-form generation.

\textbf{Literary domain (20 documents, 80 queries).} Public-domain books from authors including H.G.\ Wells, Jules Verne, Dickens, and Shelley, sourced from Internet Archive~\cite{internetarchive}. These long-form narratives contain complex characterization, thematic arcs, and non-linear plots spanning tens of thousands of tokens, stressing long-range coherence and structured narrative generation.

\textbf{Financial domain (20 documents, 40 queries).} Real-world annual reports and economic outlook documents from organizations such as Reliance Industries, TCS, Infosys, SBI, the World Bank, and Goldman Sachs. These documents are analytically dense, containing numerical data, multi-year comparisons, and domain-specific terminology, thereby stressing factual retrieval and numerical grounding.

\textbf{Legal domain (20 documents, 50 queries).} U.S.\ federal court case materials from the Multi-LexSum dataset~\cite{shen2022multilexsum}, involving complex legal reasoning, multi-party argumentation, and highly formalized language. This domain primarily stresses factual consistency and domain-specific reasoning reliability.

Queries were manually authored by annotators to simulate realistic information-seeking scenarios, averaging approximately five queries per document. Both broad prompts (e.g., ``Summarize the key arguments'') and document-specific prompts targeting entities, events, and numerical details are included to ensure varying retrieval and reasoning difficulty.

\subsection{Implementation Details}

Each RH-RAG agent---Planner, Writer, and Revision---can be instantiated with any instruction-tuned backbone. We evaluate three open-source models: \textbf{Mistral-7B-Instruct}~\cite{jiang2023mistral}, \textbf{LLaMA-3.1-8B-Instruct}~\cite{meta2024llama3}, and \textbf{Qwen-2.5-7B-Instruct}~\cite{qwen2025qwen25}, to test whether RH-RAG generalizes across architectures. All models are deployed locally on two NVIDIA T4 GPUs (16 GB VRAM each), representing accessible consumer-grade hardware.

Document preprocessing uses \texttt{bge-base-en-v1.5} for Level-2 dense encoding, while a Mistral-7B-Instruct model fine-tuned on XSum is used for Level-1 summary generation and bounded coherence memory updates. The Checker employs a pretrained DeBERTa-v3-large model for textual entailment classification. Level-1 segments are set to 20{,}000--30{,}000 tokens, and Level-2 retrieval uses top-$k{=}5$ chunks per query. The attestation loop is capped at $T{=}3$ iterations with a default coverage threshold of $\tau{=}0.8$ (additional analysis in Table~\ref{tab:threshold} and Appendix~\ref{app:ablation}).

Each agent uses a dedicated role-specific prompt (Appendix~\ref{app:prompts}): planning prompts generate structured outlines with retrieval anchors, writing prompts condition on evidence and coherence memory, and revision prompts focus on correcting unsupported claims while preserving structure.

\subsection{Evaluation Metrics}

We evaluate systems across output quality and factual faithfulness.

\textbf{Output quality (Win Rate).} We adopt an LLM-as-a-judge framework~\cite{zheng2023judging} using GPT-4o as an external evaluator to avoid self-preference bias. RH-RAG outputs and baseline responses are compared pairwise on coherence, relevance, and overall writing quality. Each comparison is evaluated twice with swapped ordering to reduce positional bias; the final win rate reflects the proportion of preferred outputs.

\textbf{Faithfulness metrics.} \textbf{SUMMAC}~\cite{laban2021summac} measures entailment-based semantic consistency between generated text and source documents using sentence-level NLI aggregation. \textbf{AlignScore}~\cite{zha2023alignscore} evaluates broader information alignment through chunk-aware text-to-text alignment. Together, these metrics capture complementary aspects of factual grounding and semantic fidelity.

\subsection{Baselines}

We compare RH-RAG against four baseline categories.

\textbf{Standard RAG}~\cite{lewis2020rag} uses a single-pass generation pipeline with hybrid BM25~\cite{robertson2009bm25}+dense retrieval over fixed-size chunks and Mistral-7B as the generator. It contains no planning, coherence memory, or verification, representing the default local deployment setup.

\textbf{Hierarchical RAG}~\cite{edge2024graphrag,song2024hierarchicalcontext} introduces two-level retrieval with section summaries and fine-grained chunk retrieval, but still performs single-pass generation. This baseline isolates retrieval improvements from RH-RAG's planning and verification components.

\textbf{Commercial LLMs} (GPT-4o, Claude, Gemini Pro) directly generate responses from prompts and retrieved context without the RH-RAG pipeline. These serve as upper-bound reference systems representing the performance of large proprietary models.

\textbf{RH-RAG (w/o Checker)} retains the dual-level retrieval, Planner Agent, and Writer Agent with coherence memory, but removes the Checker Agent and attestation-driven revision loop. This ablation isolates the contribution of factual verification from the remaining architectural components.
%%
%% SECTION 5: Results and Discussion
%%
\section{Results and Discussion}

\paragraph{Overall performance against retrieval baselines.}
RH-RAG outperforms both retrieval baselines across all three domains and all three metrics. On the literary domain, RH-RAG (Mistral-7B) achieves SUMMAC 47.6 / AlignScore 76.7 compared to 40.9 / 72.9 for Standard RAG and 46.3 / 71.3 for Hierarchical RAG. Notably, while Hierarchical RAG achieves a higher SUMMAC than Standard RAG (reflecting better retrieval precision), its AlignScore is actually lower in the literary domain, suggesting that more structured retrieval alone can increase thematic relevance at the expense of fine-grained information alignment. The further improvement from Hierarchical RAG to full RH-RAG isolates the contribution of structured planning and attestation-driven revision over and above retrieval architecture. Consistent improvements across financial and legal domains confirm that these gains are domain-agnostic and reflect general improvements in evidence utilization.

\paragraph{Contribution of the Checker Agent.}
Removing the attestation verification loop consistently and substantially degrades performance across all domains. On the financial domain---where numerical precision is most critical---AlignScore drops from 78.5 to 74.2 and SUMMAC from 48.7 to 42.8 when the Checker Agent is removed. Comparable degradations are observed in the literary (AlignScore $-$3.4) and legal (AlignScore $-$4.6) domains. The smaller win-rate gap between the full and ablated systems is expected, since the Checker Agent primarily improves factual grounding rather than surface fluency, and win-rate judges also weight stylistic quality.

\paragraph{Threshold sensitivity.}
Table~\ref{tab:threshold} shows that stricter coverage thresholds monotonically improve both faithfulness metrics, but with diminishing marginal returns above $\tau{=}0.8$. At $\tau{=}0.9$, stricter enforcement increases revision iterations, and in the legal domain, domain-specific phrasing absent from retrieved passages causes the NLI model to reject valid claims, triggering unnecessary revision cycles that can degrade fluency (see Appendix~\ref{app:ablation}).

\paragraph{Comparison with proprietary language models.}
RH-RAG with open-source backbones approaches proprietary systems on faithfulness metrics despite operating at a fraction of the model capacity and entirely without external data transmission. RH-RAG (Mistral-7B) achieves AlignScore 76.7 on the literary domain, within 3 points of standalone GPT-4o (79.1) and Claude (78.7)---neither of which employs a dedicated verification pipeline. The win-rate gap between RH-RAG variants (6.7--7.2) and commercial baselines (7.8--8.4) reflects differences in surface fluency and instruction-following breadth, which are properties of the underlying backbone capacity rather than the RH-RAG framework itself. A representative full-length output from RH-RAG is provided in Appendix~\ref{app:sample_output}.

%%
%% SECTION 7: Conclusion
%%
\section{Conclusion}

We presented RH-RAG, a multi-agent RAG framework combining dual-level retrieval, structured section-wise generation with bounded coherence memory, and attestation-driven NLI-based factual verification for privacy-constrained long-form generation. RH-RAG substantially outperforms standard and hierarchical RAG baselines and approaches the faithfulness of proprietary cloud systems, all without transmitting sensitive data to external providers. Future directions include domain-adaptive verification thresholds, agent-specific fine-tuning, and more efficient specialized NLI models to reduce verification overhead in technical domains.

%%
%% SECTION 6: Limitations
%%
\section{Limitations}

Three limitations of the current system are worth noting. First, the framework is deliberately constrained to 7B--8B models, meaning that highly complex multi-step logical deductions and intricate narrative transitions can still challenge the underlying backbone, occasionally increasing revision cycles. Second, all pipeline components use general-purpose instruction-tuned models; agent-specific fine-tuning---particularly a domain-adapted NLI model for the Checker Agent---would reduce unnecessary revisions and lower computational overhead. Third, our literary evaluation relies on well-known public-domain texts whose content likely appears in the pre-training corpora of the evaluated models. While the verification metric strictly measures grounding against retrieved evidence, latent parametric familiarity with these narratives may artificially facilitate coherence, representing a potential upper bound on performance relative to fully novel confidential documents.

\end{multicols}
%%
%% Bibliography
%%
\bibliographystyle{unsrt}
\nocite{*}
\bibliography{references}

@article{andryushchenko2024llmcodeqa,
  title={Leveraging Large Language Models in Code Question Answering: Baselines and Issues},
  author={Andryushchenko, Georgy and Ivanov, Vladimir and Makharev, Vladimir and Tukhtina, Elizaveta and Valeev, Aidar},
  journal={arXiv preprint arXiv:2411.03012},
  year={2024}
}

@article{asai2024reliable,
  title={Reliable, Adaptable, and Attributable Language Models with Retrieval},
  author={Asai, Akari and Zhong, Zexuan and Chen, Danqi and Koh, Pang Wei and Zettlemoyer, Luke and Hajishirzi, Hannaneh and Yih, Wen-tau},
  journal={arXiv preprint arXiv:2403.03187},
  year={2024}
}

@inproceedings{shao2024storm,
  title={Assisting in Writing Wikipedia-like Articles From Scratch with Large Language Models},
  author={Shao, Yijia and Jiang, Yucheng and Kanell, Theodore A. and Xu, Peter and Khattab, Omar and Lam, Monica S.},
  booktitle={Proceedings of the 2024 Conference of the North American Chapter of the Association for Computational Linguistics (NAACL)},
  year={2024},
  address={Mexico City, Mexico},
  publisher={Association for Computational Linguistics},
  pages={6252--6278},
  doi={10.18653/v1/2024.naacl-long.347}
}

@article{beltagy2020longformer,
  title={Longformer: The Long-Document Transformer},
  author={Beltagy, Iz and Peters, Matthew E. and Cohan, Arman},
  journal={arXiv preprint arXiv:2004.05150},
  year={2020}
}

@article{bertsch2023unlimiformer,
  title={Unlimiformer: Long-Range Transformers with Unlimited Length Input},
  author={Bertsch, Amanda and Alon, Uri and Neubig, Graham and Gormley, Matthew R.},
  journal={arXiv preprint arXiv:2305.01625},
  year={2023}
}

@article{bian2023gosum,
  title={GoSum: Extractive Summarization of Long Documents by Reinforcement Learning and Graph Organized Discourse State},
  author={Bian, Junyi and Huang, Xiaodi and Zhou, Hong and Zhu, Shanfeng},
  journal={arXiv preprint arXiv:2211.10247},
  year={2023}
}

@inproceedings{bohnet2022attributedqa,
  title={Attributed Question Answering: Evaluation and Modeling for Attributed Large Language Models},
  author={Bohnet, Bernd and Tran, Vinh and Verga, Pat and Aharoni, Roee and others},
  booktitle={Proceedings of EMNLP},
  year={2022}
}

@inproceedings{chang2024booookscore,
  title={BooookScore: A Systematic Exploration of Book-Length Summarization in the Era of LLMs},
  author={Chang, Yapei and Lo, Kyle and Goyal, Tanya and Iyyer, Mohit},
  booktitle={International Conference on Learning Representations},
  year={2024}
}

@article{chen2023positional,
  title={Extending Context Window of Large Language Models via Positional Interpolation},
  author={Chen, Shouyuan and Wong, Sherman and Chen, Liangjian and Tian, Yuandong},
  journal={arXiv preprint arXiv:2306.15595},
  year={2023}
}

@article{chen2024longlora,
  title={LongLoRA: Efficient Fine-Tuning of Long-Context Large Language Models},
  author={Chen, Yukang and Qian, Shengju and Tang, Haotian and Lai, Xin and Liu, Zhijian and Han, Song and Jia, Jiaya},
  journal={arXiv preprint arXiv:2309.12307},
  year={2024}
}

@article{child2019sparse,
  title={Generating Long Sequences with Sparse Transformers},
  author={Child, Rewon and Gray, Scott and Radford, Alec and Sutskever, Ilya},
  journal={arXiv preprint arXiv:1904.10509},
  year={2019}
}

@inproceedings{cohan2018discourse,
  title={A Discourse-Aware Attention Model for Abstractive Summarization of Long Documents},
  author={Cohan, Arman and Dernoncourt, Franck and Kim, Doo Soon and others},
  booktitle={NAACL-HLT},
  year={2018}
}

@article{fierro2024planningcitations,
  title={Learning to Plan and Generate Text with Citations},
  author={Fierro, Constanza and Amplayo, Reinald Kim and Huot, Fantine and De Cao, Nicola and Maynez, Joshua and Narayan, Shashi and Lapata, Mirella},
  journal={arXiv preprint arXiv:2404.03381},
  year={2024}
}

@article{frantar2023gptq,
  title={GPTQ: Accurate Post-Training Quantization for Generative Pre-trained Transformers},
  author={Frantar, Elias and Ashkboos, Saleh and Hoefler, Torsten and Alistarh, Dan},
  journal={arXiv preprint arXiv:2210.17323},
  year={2023}
}

@inproceedings{gao2023citations,
  title={Enabling Large Language Models to Generate Text with Citations},
  author={Gao, Tianyu and Yen, Howard and Yu, Jiatong and Chen, Danqi},
  booktitle={EMNLP},
  year={2023}
}

@inproceedings{gu2022memsum,
  title={MemSum: Extractive Summarization of Long Documents Using Multi-step Episodic Markov Decision Processes},
  author={Gu, Nianlong and Ash, Elliott and Hahnloser, Richard},
  booktitle={ACL},
  year={2022}
}

@article{he2021deberta,
  title={DeBERTa: Decoding-Enhanced BERT with Disentangled Attention},
  author={He, Pengcheng and Liu, Xiaodong and Gao, Jianfeng and Chen, Weizhu},
  journal={arXiv preprint arXiv:2006.03654},
  year={2021}
}

@inproceedings{izacard2021leveraging,
  title={Leveraging Passage Retrieval with Generative Models for Open Domain Question Answering},
  author={Izacard, Gautier and Grave, Edouard},
  booktitle={EACL},
  year={2021}
}

@article{katharopoulos2020transformersrnn,
  title={Transformers Are RNNs: Fast Autoregressive Transformers with Linear Attention},
  author={Katharopoulos, Angelos and Vyas, Apoorv and Pappas, Nikolaos and Fleuret, François},
  journal={ICML},
  year={2020}
}

@inproceedings{kryscinski2022booksum,
  title={BOOKSUM: A Collection of Datasets for Long-Form Narrative Summarization},
  author={Kryscinski, Wojciech and Rajani, Nazneen and Agarwal, Divyansh and Xiong, Caiming and Radev, Dragomir},
  booktitle={Findings of EMNLP},
  year={2022}
}

@article{li2024longcontext,
  title={Long-Context LLMs Struggle with Long In-Context Learning},
  author={Li, Tianle and Zhang, Ge and Do, Quy Duc and Yue, Xiang and Chen, Wenhu},
  journal={arXiv preprint arXiv:2404.02060},
  year={2024}
}

@inproceedings{lin2004rouge,
  title={ROUGE: A Package for Automatic Evaluation of Summaries},
  author={Lin, Chin-Yew},
  booktitle={ACL Workshop on Text Summarization},
  year={2004}
}

@inproceedings{liu2019bertSum,
  title={Text Summarization with Pretrained Encoders},
  author={Liu, Yang and Lapata, Mirella},
  booktitle={EMNLP},
  year={2019}
}

@inproceedings{min2023factscore,
  title={FActScore: Fine-Grained Atomic Evaluation of Factual Precision in Long Form Text Generation},
  author={Min, Sewon and Krishna, Kalpesh and others},
  booktitle={EMNLP},
  year={2023}
}

@inproceedings{shaham2023zeroscrolls,
  title={ZeroSCROLLS: A Zero-Shot Benchmark for Long Text Understanding},
  author={Shaham, Uri and Ivgi, Maor and Efrat, Avia and Berant, Jonathan and Levy, Omer},
  booktitle={EMNLP Findings},
  year={2023}
}

@article{shi2023replug,
  title={RePlug: Retrieval-Augmented Black-Box Language Models},
  author={Shi, Weijia and Min, Sewon and Yasunaga, Michihiro and others},
  journal={arXiv preprint arXiv:2301.12652},
  year={2023}
}

@article{song2024hierarchicalcontext,
  title={Hierarchical Context Merging: Better Long Context Understanding for Pre-trained LLMs},
  author={Song, Woomin and Oh, Seunghyuk and Mo, Sangwoo and others},
  journal={arXiv preprint arXiv:2404.10308},
  year={2024}
}

@inproceedings{zhang2023extractiveunfaithful,
  title={Extractive is Not Faithful: An Investigation of Broad Unfaithfulness Problems in Extractive Summarization},
  author={Zhang, Shiyue and Wan, David and Bansal, Mohit},
  booktitle={ACL},
  year={2023}
}

@misc{internetarchive,
  author = {{Internet Archive}},
  title = {Wayback Machine},
  year = {2024},
  howpublished = {\url{https://archive.org/web/}},
  note = {Accessed: 2026-03-16}
}

@article{shen2022multilexsum,
  title={Multi-LexSum: Real-World Summaries of Civil Rights Lawsuits at Multiple Granularities},
  author={Shen, Zejiang and Lo, Kyle and Yu, Lauren and Dahlberg, Nathan and Schlanger, Margo and Downey, Doug},
  journal={arXiv preprint arXiv:2206.10883},
  year={2022}
}

@article{meta2024llama3,
  title={The Llama 3 Herd of Models},
  author={{Meta AI}},
  journal={arXiv preprint arXiv:2407.21783},
  year={2024}
}

@article{qwen2025qwen25,
  title={Qwen2.5 Technical Report},
  author={{Qwen Team}},
  journal={arXiv preprint arXiv:2412.15115},
  year={2025}
}

@article{jiang2023mistral,
  title={Mistral 7B},
  author={Jiang, Albert Q. and Sablayrolles, Alexandre and Roux, Antoine and Mensch, Arthur and others},
  journal={arXiv preprint arXiv:2310.06825},
  year={2023}
}

@inproceedings{zheng2023judging,
  title={Judging LLM-as-a-Judge with MT-Bench and Chatbot Arena},
  author={Zheng, Lianmin and Chiang, Wei-Lin and Sheng, Ying and Zhuang, Siyuan and Wu, Zhanghao and Zhuang, Yonghao and Lin, Zi and Li, Zhuohan and Li, Dacheng and Xing, Eric P. and Stoica, Ion and Gonzalez, Joseph E.},
  booktitle={Advances in Neural Information Processing Systems},
  year={2023}
}

@article{bai2023lmevaluator,
  title={Benchmarking Foundation Models with Language-Model-as-an-Examiner},
  author={Bai, Yushi and Ying, Jiahao and Cao, Yixin and Lv, Xin and He, Yuze and others},
  journal={Advances in Neural Information Processing Systems},
  year={2023}
}

@article{zheng2023lmsyschat,
  title={LMSYS-Chat-1M: A Large-Scale Real-World LLM Conversation Dataset},
  author={Zheng, Lianmin and Chiang, Wei-Lin and Sheng, Ying and others},
  journal={arXiv preprint arXiv:2309.11998},
  year={2023}
}

@article{laban2021summac,
  title={SummaC: Re-visiting NLI-based Models for Inconsistency Detection in Summarization},
  author={Laban, Philippe and Schnabel, Tobias and Bennett, Paul N. and Hearst, Marti A.},
  journal={arXiv preprint arXiv:2111.09525},
  year={2021}
}

@inproceedings{zha2023alignscore,
  title={AlignScore: Evaluating Factual Consistency with A Unified Alignment Function},
  author={Zha, Yuheng and Yang, Yichi and Li, Ruichen and Hu, Zhiting},
  booktitle={Proceedings of the 61st Annual Meeting of the Association for Computational Linguistics (ACL)},
  year={2023}
}

@inproceedings{lewis2020rag,
  title={Retrieval-Augmented Generation for Knowledge-Intensive NLP Tasks},
  author={Lewis, Patrick and Perez, Ethan and Piktus, Aleksandra and Petroni, Fabio and Karpukhin, Vladimir and Goyal, Naman and Küttler, Heinrich and Lewis, Mike and Yih, Wen-tau and Rocktäschel, Tim and Riedel, Sebastian and Kiela, Douwe},
  booktitle={Advances in Neural Information Processing Systems},
  year={2020}
}

@article{robertson2009bm25,
  title={The Probabilistic Relevance Framework: BM25 and Beyond},
  author={Robertson, Stephen and Zaragoza, Hugo},
  journal={Foundations and Trends in Information Retrieval},
  year={2009}
}

@article{edge2024graphrag,
  title={From Local to Global: A Graph RAG Approach to Query-Focused Summarization},
  author={Edge, Darren and Trinh, Ha and Cheng, Newman and Bradley, Joshua and Chao, Alex and Mody, Apurva and Truitt, Steven and Larson, Jonathan},
  journal={arXiv preprint arXiv:2404.16130},
  year={2024}
}

@article{asai2023selfrag,
  title={Self-RAG: Learning to Retrieve, Generate, and Critique Through Self-Reflection},
  author={Asai, Akari and Wu, Zeqiu and Wang, Yizhong and Sil, Avirup and Hajishirzi, Hannaneh},
  journal={arXiv preprint arXiv:2310.11511},
  year={2023}
}

@article{mirage2026scaling,
  title={MIRAGE: Scaling Test-Time Inference with Parallel Graph-Retrieval-Augmented Reasoning Chains},
  author={Kaiwen and Wei, Rui and Shan, Dongsheng and Zou, Jianzhong and Yang, Bi and Zhao, Junnan and Zhu, Jiang and Zhong},
  journal={arXiv preprint arXiv:2508.18260},
  year={2025}
}

@inproceedings{wen2024mindmap,
  title={MindMap: Knowledge Graph Prompting Sparks Graph of Thoughts in Large Language Models},
  author={Wen, Yilin and Wang, Zifeng and Sun, Jimeng},
  booktitle={Proceedings of the 62nd Annual Meeting of the Association for Computational Linguistics (Volume 1: Long Papers)},
  pages={10370--10388},
  year={2024},
  publisher={Association for Computational Linguistics}
}

%%
%% APPENDIX
%%
\appendix

\begin{multicols}{2}

%%
%% APPENDIX A: Prompt Templates
%%
\section{Prompt Templates for RH-RAG Agents}
\label{app:prompts}

This appendix provides the full prompt templates for each agent in the RH-RAG pipeline. Each agent operates with a dedicated role-conditioned prompt injected with structured context at inference time. The three agents---Planner, Writer, and Rewriter---perform structurally distinct functions and accordingly receive different contextual inputs.

\subsection{Planner Agent Prompt}

The Planner Agent analyzes the user query alongside Level-1 routing summaries to produce a structured generation plan before any text is written. Its three core functions are: (i) identifying the main objective of the query; (ii) constructing a logically ordered multi-section outline; and (iii) generating per-section retrieval anchors that link the high-level plan to Level-2 evidence chunks.

\medskip
\noindent\textbf{Planner Prompt:}

\smallskip
\noindent\textit{``You are a document planning system.}

\medskip
\noindent\textit{Task:}

\noindent\textit{Create a structured generation plan for a long-form document.}

\medskip
\noindent\textit{Instructions:}

\noindent\textit{1. Understand the query and determine:}

\noindent\textit{\phantom{xxx} * objective}

\noindent\textit{\phantom{xxx} * document type}

\noindent\textit{\phantom{xxx} * required scope}

\medskip
\noindent\textit{2. Analyze routing summaries and identify:}

\noindent\textit{\phantom{xxx} * major themes}

\noindent\textit{\phantom{xxx} * important concepts}

\noindent\textit{\phantom{xxx} * required evidence areas}

\medskip
\noindent\textit{3. Divide the document into logically ordered sections.}

\medskip
\noindent\textit{Section Rules:}

\noindent\textit{\phantom{xxx} * Each section should focus on one major idea}

\noindent\textit{\phantom{xxx} * Avoid redundancy and overlapping sections}

\noindent\textit{\phantom{xxx} * Ensure sections collectively cover the topic}

\noindent\textit{\phantom{xxx} * Order sections logically}

\medskip
\noindent\textit{For each section generate:}

\noindent\textit{\phantom{xxx} * section\_title: concise title}

\noindent\textit{\phantom{xxx} * section\_description: short explanation of what the section covers and why it exists}

\noindent\textit{\phantom{xxx} * retrieval\_anchor: list of keywords or short phrases useful for retrieval}

\medskip
\noindent\textit{Retrieval Anchor Rules:}

\noindent\textit{\phantom{xxx} * keywords / short phrases only}

\noindent\textit{\phantom{xxx} * avoid generic terms}

\noindent\textit{\phantom{xxx} * prefer entities, concepts, methods, terminology}

\medskip
\noindent\textit{Output Rules:}

\noindent\textit{\phantom{xxx} * Output ONLY the plan}

\noindent\textit{\phantom{xxx} * No reasoning}

\noindent\textit{\phantom{xxx} * No explanations}

\medskip
\noindent\textit{Output Format:}

\noindent\textit{[\{}

\noindent\textit{\phantom{xxx} ``section\_title'': ``...'',}

\noindent\textit{\phantom{xxx} ``section\_description'': ``...'',}

\noindent\textit{\phantom{xxx} ``retrieval\_anchor'': [``...'', ``...'']}

\noindent\textit{\}]''}

\subsection{Writer Agent Prompt}

The Writer Agent generates document content section by section, conditioned on the structured outline, Level-2 retrieved evidence, and the bounded coherence memory from prior sections. It is instructed to avoid introducing unsupported information and to maintain narrative continuity.

\medskip
\noindent\textbf{Writer Prompt:}

\smallskip
\noindent\textit{``You are a technical writing system responsible for generating ONE section of a larger document.}

\medskip
\noindent\textit{Task:}

\noindent\textit{Generate the complete content for the current section.}

\medskip
\noindent\textit{Rules:}

\noindent\textit{\phantom{xxx} * Focus ONLY on the current section objective}

\noindent\textit{\phantom{xxx} * Treat this section as ONE component of a larger document}

\noindent\textit{\phantom{xxx} * Maintain continuity with previous sections without repeating them}

\noindent\textit{\phantom{xxx} * Use retrieved evidence to support claims}

\noindent\textit{\phantom{xxx} * Do not introduce unsupported information}

\noindent\textit{\phantom{xxx} * If evidence is limited, remain concise rather than hallucinating}

\medskip
\noindent\textit{Structure Requirements:}

\noindent\textit{\phantom{xxx} * Use the section title as the heading}

\noindent\textit{\phantom{xxx} * Produce a balanced section length}

\noindent\textit{\phantom{xxx} * Aim for approximately similar depth and detail as other sections}

\noindent\textit{\phantom{xxx} * Develop multiple ideas when evidence permits rather than stopping early}

\noindent\textit{\phantom{xxx} * Use multiple paragraphs with logical progression}

\noindent\textit{\phantom{xxx} * Do NOT generate introductions for the entire document}

\noindent\textit{\phantom{xxx} * Do NOT generate conclusions, summaries, or closing remarks unless explicitly required by the section description}

\noindent\textit{\phantom{xxx} * Do NOT discuss future sections}

\medskip
\noindent\textit{Style Requirements:}

\noindent\textit{\phantom{xxx} * Clear and information dense}

\noindent\textit{\phantom{xxx} * Avoid filler and repetition}

\noindent\textit{\phantom{xxx} * Write naturally as part of a continuous document}

\noindent\textit{\phantom{xxx} * Do not mention retrieval, evidence, memories, or summaries}

\medskip
\noindent\textit{Output ONLY the section text''}

\subsection{Rewriter Agent Prompt}

The Rewriter Agent is invoked when a section fails the coverage threshold. It receives the original section, a list of unsupported or contradicted claims identified by the Checker Agent, and newly retrieved evidence passages. The agent corrects only the failed claims while preserving the surrounding paragraph structure and meaning.

\medskip
\noindent\textbf{Rewriter Prompt:}

\smallskip
\noindent\textit{``You are a factual revision system.}

\medskip
\noindent\textit{Task:}

\noindent\textit{Revise the section so that unsupported statements are corrected, replaced, or removed using the provided evidence.}

\medskip
\noindent\textit{Rules:}

\noindent\textit{\phantom{xxx} * Modify ONLY statements related to unsupported claims}

\noindent\textit{\phantom{xxx} * Preserve original wording, structure, flow, and style whenever possible}

\noindent\textit{\phantom{xxx} * Use retrieved evidence to support revisions}

\noindent\textit{\phantom{xxx} * Do not introduce unsupported information}

\noindent\textit{\phantom{xxx} * If evidence is insufficient, remove or weaken unsupported claims rather than inventing facts}

\noindent\textit{\phantom{xxx} * Avoid unnecessary rewriting}

\noindent\textit{\phantom{xxx} * Maintain coherence after edits}

\medskip
\noindent\textit{Output Requirements:}

\noindent\textit{\phantom{xxx} * Return the complete revised section}

\noindent\textit{\phantom{xxx} * Do not explain changes}

\noindent\textit{\phantom{xxx} * Do not output reasoning}

\noindent\textit{\phantom{xxx} * Output ONLY the revised section''}

%%
%% APPENDIX B: Dataset Details
%%
\section{Dataset Details}
\label{app:dataset}

The evaluation dataset covers three distinct document domains to assess RH-RAG across narrative understanding, factual grounding, and domain-specific reasoning.

\subsection{Literary Documents}

Public-domain books spanning science fiction, classic literature, and philosophy, sourced from Internet Archive~\cite{internetarchive}. These test long-range coherence, narrative structure, and thematic understanding. Titles include:
\textit{The Time Machine}, \textit{The Invisible Man}, \textit{The War of the Worlds} (H.G.\ Wells); \textit{Flatland} (Abbott); \textit{Twenty Thousand Leagues Under the Sea}, \textit{Journey to the Center of the Earth}, \textit{From the Earth to the Moon} (Jules Verne); \textit{Frankenstein} (Shelley); \textit{The Picture of Dorian Gray} (Wilde); \textit{The Turn of the Screw} (James); \textit{Great Expectations}, \textit{David Copperfield} (Dickens); \textit{Gulliver's Travels} (Swift); \textit{Alice's Adventures in Wonderland} (Carroll); \textit{Meditations} (Aurelius); and \textit{The Republic} (Plato).

\textbf{Note on potential data contamination:} Because these are widely distributed pre-training texts, the underlying LLMs may possess latent parametric knowledge of their narratives. While our verification metric strictly measures grounding against retrieved external evidence, this prior familiarity could facilitate narrative coherence. Performance on these public texts may therefore represent an upper bound relative to entirely novel confidential documents.

\subsection{Financial Documents}

Real-world annual and economic reports with dense numerical data, structured analysis, and domain-specific terminology, used to evaluate precise factual retrieval and numerical grounding. Documents include:
Lenskart Annual Report 2024, Reliance Industries Annual Report 2024, TCS Annual Report 2024, Infosys Annual Report 2024, SBI Annual Report 2024, World Bank Economic Report, Goldman Sachs Outlook Report, and the Lenskart IPO Report.

\subsection{Legal Documents}

Documents from the Multi-LexSum dataset~\cite{shen2022multilexsum}, comprising U.S.\ federal court case materials with highly structured argumentation, legal judgments, and case details. These evaluate domain-specific reasoning under strict factual consistency requirements.

\subsection{Query Construction}

Queries include both \textbf{general queries} applicable across multiple documents (e.g., ``Summarize the key ideas presented in the document,'' ``Identify the main themes or arguments,'' ``Extract the most important factual information'') and \textbf{document-specific queries} targeting particular sections or topics. Example document-specific queries for \textit{The Invisible Man}: ``Explain the character of Griffin, the Invisible Man'' and ``Explain in detail the plan of enforcement of the Reign of Terror by Griffin.''

%%
%% APPENDIX C: Extended Ablation Study
%%
\section{Extended Ablation Study}
\label{app:ablation}

\subsection{Contribution of Individual Pipeline Stages}

To understand the marginal contribution of each component, we compare four progressive configurations on the financial domain using the Mistral-7B backbone:

\begin{itemize}
    \item \textbf{Hierarchical RAG:} two-level retrieval, single-pass generation.
    \item \textbf{+ Planner Agent:} adds structured outline generation before writing.
    \item \textbf{+ Writer Agent (bounded memory):} adds section-wise generation with coherence memory.
    \item \textbf{Full RH-RAG:} adds Checker Agent with attestation feedback loop.
\end{itemize}

Each stage provides incremental gains in SUMMAC and AlignScore. The Planner Agent contributes the largest improvement to win rate (improving document-level structural coherence), while the Checker Agent yields the largest individual gain in AlignScore (approximately +4.3 points over the Writer-only ablation), confirming the non-redundant contribution of each architectural component.

\subsection{Threshold Sensitivity and Revision Overhead}

As reported in Table~\ref{tab:threshold} of the main paper, increasing $\tau$ monotonically improves both SUMMAC and AlignScore, but marginal gains diminish above $\tau{=}0.8$ while revision iterations increase. At $\tau{=}0.9$, domain-specific terminology (particularly in legal documents) causes the NLI model to reject valid claims, leading to unnecessary revision cycles that occasionally degrade fluency without improving factual accuracy. The $\tau{=}0.8$ operating point provides the best faithfulness--efficiency trade-off across all three domains.

These observations motivate \textbf{domain-adaptive threshold scheduling}: applying a lower threshold for narrative domains (where NLI reliability is higher) and a stricter threshold for technical domains (where evidence alignment is more critical). This represents a concrete direction for future work.

\subsection{Impact of Coherence Memory Design}

We ablate the bounded coherence memory by replacing it with: (i)~no prior context, (ii)~the full previously generated text (unbounded), and (iii)~the proposed dual-component memory (local $+$ global summary). The unbounded setting degrades performance on longer documents due to context saturation exceeding the effective attention window of the 7B model. The no-context setting leads to progressive narrative drift between sections, with characters, arguments, or established facts from early sections being forgotten or contradicted. The dual-component memory achieves the best balance, maintaining both local and global continuity without exceeding practical context limits.

%%
%% APPENDIX D: Sample Output
%%
\section{Sample Output}
\label{app:sample_output}

\noindent\textbf{Reference document:} \textit{Rich Dad and Poor Dad} by Robert Kiyosaki\\
\noindent\textbf{Prompt:} how to become financially independent?\\
\noindent\textbf{LLM Used:} Mistral-7B-Instruct-v0.1\\

\medskip
\noindent\textbf{Output:}

\medskip
\noindent\textbf{A Comprehensive Guide to Becoming Financially Independent}

\medskip
\noindent\textbf{Introduction}\\
Becoming financially independent is a goal that many people strive for, but few achieve. It requires a combination of financial education, discipline, and smart investment strategies. In this document, we will provide a comprehensive guide to becoming financially independent, covering essential topics such as cash flow, people, and personal time.

\medskip
Cash flow is the lifeblood of any business or individual. It is essential to manage your cash flow effectively to ensure that you have enough money coming in to cover your expenses. This includes paying yourself first, which is a crucial habit to develop. By prioritizing your own financial needs, you can ensure that you are always in a good financial position.

\medskip
People are another important aspect of becoming financially independent. Surrounding yourself with the right people can make all the difference in achieving your financial goals. This includes mentors, advisors, and colleagues who can provide guidance and support.

\medskip
Personal time is also essential for achieving financial independence. It is important to take time for yourself to relax, recharge, and pursue your passions. This can help you stay motivated and focused on your financial goals.

\medskip
Mastering self-discipline is key to achieving financial independence. This includes developing good habits such as saving, investing, and managing your cash flow effectively. It also involves being disciplined with your time and energy, as well as your emotions and thoughts.

\medskip
In conclusion, becoming financially independent requires a combination of financial education, discipline, and smart investment strategies. By focusing on cash flow, people, and personal time, and mastering self-discipline, you can achieve your financial goals and achieve financial independence.

\medskip
\noindent\textbf{Financial Education}\\
Financial education is a crucial aspect of becoming financially independent. It is often overlooked in traditional education, which tends to focus solely on physical, emotional, and spiritual growth. However, financial genius exists but is suppressed by cultural education.

\medskip
To achieve financial independence, it is essential to understand the importance of cash flow, people, and personal time. Prioritizing your own financial needs is crucial for effective cash flow management. Surrounding yourself with the right people can provide guidance and support. Taking time for yourself is important for achieving financial independence and staying motivated.

\medskip
Mastering self-discipline is key to achieving financial independence and financial goals. It requires good habits and emotional control. Without emotional and spiritual development, it is challenging to make a permanent change in life. For example, it rarely works to tell an overweight person, ``Just eat less and exercise more.'' Diet and exercise may make sense mentally, but most people who are overweight do not eat because they are hungry. They eat to feed an emptiness in their emotions and their soul.

\medskip
To become financially independent, it is important to have a long memory of the document's comprehensive guide to achieving financial independence. Cash flow, people, and personal time are essential topics that will be covered. Prioritizing your own financial needs is crucial for effective cash flow management. Surrounding yourself with the right people can provide guidance and support. Taking time for yourself is important for achieving financial independence and staying motivated.

\medskip
In contrast, a short memory of the document's guide to becoming financially independent is not enough. Cash flow is essential for financial independence. Prioritizing your own financial needs is crucial for effective cash flow management. Surrounding yourself with the right people can provide guidance and support. Taking time for yourself to relax and pursue passions is important for achieving financial independence and staying motivated. Mastering self-discipline through good habits and emotional control is key to achieving financial independence and financial goals.

\medskip
\noindent\textbf{Paying Yourself First}\\
Paying yourself first is a financial strategy that can lead to financial freedom and success. By prioritizing your own financial needs, you can avoid creditor control and increase your motivation to work harder and smarter.

\medskip
One of the benefits of paying yourself first is financial motivation. When you prioritize your own financial needs, you create a sense of urgency to work harder and earn more money. This pressure can force you to seek out other forms of income, such as starting a side business or investing in the stock market.

\medskip
Another benefit of paying yourself first is avoiding creditor control. When you pay yourself first, you reduce the amount of money you owe to creditors. This can help you avoid high-interest rates and other penalties, which can make it harder to get out of debt.

\medskip
Paying yourself first can also lead to financial freedom. By prioritizing your own financial needs, you can create a plan to achieve your financial goals, such as buying a home or starting a business. This can help you break free from the cycle of debt and build a more secure financial future.

\medskip
To implement the pay yourself first strategy, you should start by setting a financial goal. This could be a specific amount of money you want to save or a specific financial goal you want to achieve, such as buying a home or starting a business.

\medskip
Next, you should create a budget that prioritizes your own financial needs. This means setting aside a portion of your income each month for savings and investments, before paying for anything else.

\medskip
Finally, you should monitor your progress and adjust your budget as needed. This means regularly reviewing your financial situation and making adjustments to your budget to ensure you are on track to achieve your financial goals.

\medskip
In conclusion, paying yourself first is a powerful financial strategy that can lead to financial freedom and success. By prioritizing your own financial needs, you can avoid creditor control, increase your motivation to work harder and earn more money, and create a plan to achieve your financial goals. By implementing the pay yourself first strategy, you can break free from the cycle of debt and build a more secure financial future.

\medskip
\noindent\textbf{The CASHFLOW Quadrant}\\
The CASHFLOW Quadrant is a tool used to help individuals understand their financial situation and make informed decisions. It divides people into four quadrants based on where their cash comes from: E (Employee), S (Self-employed), B (Business Owner), and I (Investor). Employees and self-employed individuals reside on the left side of the CASHFLOW Quadrant, while those on the right side receive their cash from businesses they own or investments they own.

\medskip
Employees make their business owner or the shareholders rich, not themselves. They work for the company, and their efforts and success will help provide for the owner's success and retirement. On the other hand, self-employed individuals work for themselves and have more control over their financial situation. However, they also bear more responsibility and risk.

\medskip
Business owners and investors, on the other hand, receive their cash from businesses they own or investments they own. They have more control over their financial situation and can potentially achieve financial independence.

\medskip
To achieve financial independence, it is crucial to prioritize your own financial needs and manage your cash flow effectively. Financial education is also crucial, as it can provide guidance and support in achieving financial independence. Surrounding yourself with the right people can also provide valuable insights and support.

\medskip
Taking time for yourself is also important for achieving financial independence and staying motivated. Mastering self-discipline through good habits and emotional control is key to achieving financial independence and financial goals.

\medskip
One financial strategy that can lead to financial freedom and success is paying yourself first. This involves setting a financial goal, creating a budget that prioritizes savings and investments, and monitoring progress. By prioritizing your own financial needs, you can avoid creditor control and achieve financial freedom.

\medskip
\noindent\textbf{Income Differences and Investment Skills}\\
In order to achieve financial independence, it is important to understand the concept of income differences and investment skills. Traditional education often neglects these topics, but they can be learned through experiential learning and CASHFLOW games.

\medskip
One of the reasons why I became interested in these topics was after attending a personal-development seminar. This seminar rekindled my love of learning, but not the type of learning that is taught in school. I began to ask questions such as:

\begin{itemize}
    \item How can I learn in a hands-on environment and accelerate my learning curve?
    \item How can I maximize my potential by learning from those who practice what they preach?
    \item How can I gain exclusive access to top-tier experts?
\end{itemize}

\medskip
The more I studied, the more curious I became about traditional education. I began to see that it often neglects important topics such as income differences and investment skills.

\medskip
One of the key tools for understanding financial situation and making informed decisions is the CASHFLOW Quadrant. This tool divides people into four quadrants based on where their cash comes from: E (Employee), S (Self-employed), B (Business Owner), and I (Investor). Employees and self-employed individuals reside on the left side of the CASHFLOW Quadrant, while those on the right side receive their cash from businesses they own or investments they own.

\medskip
Financial education is crucial for achieving financial independence. Cash flow, people, and personal time are essential topics for financial independence. Prioritizing your own financial needs is crucial for effective cash flow management. Surrounding yourself with the right people can provide guidance and support. Taking time for yourself is important for achieving financial independence and staying motivated.

\medskip
One example of a synergy of skills is that of a young writer for the newspaper. If she diligently learned the skills of sales and marketing, her income would jump dramatically. If I were her, I would take some courses in advertising copywriting as well as sales. Then, instead of working at the newspaper, I would seek a job at an advertising agency. Even if it were a cut in pay, she would learn how to communicate in short-cuts that are used in successful advertising.

\medskip
In conclusion, understanding income differences and investment skills is crucial for achieving financial independence. Traditional education often neglects these topics, but they can be learned through experiential learning and CASHFLOW games. Financial education, cash flow, people, and personal time are essential topics for financial independence. Prioritizing your own financial needs, surrounding yourself with the right people, taking time for yourself, and mastering self-discipline through good habits and emotional control are key to achieving financial independence and financial goals.

\medskip
\noindent\textbf{Avoiding Consumer Debt}\\
Becoming financially independent is a goal that many people strive for, but few achieve. One of the most important steps towards achieving financial independence is avoiding consumer debt. Consumer debt can be a major obstacle to building wealth and achieving financial freedom.

\medskip
When you are in debt, you are spending more money than you are earning. This can lead to a cycle of debt that can be difficult to break. The more debt you accumulate, the more interest you will have to pay, and the longer it will take to pay off your debts.

\medskip
On the other hand, when you focus on building assets, you are investing in your future. Assets are things that generate income, such as stocks, real estate, and businesses. When you have a diversified portfolio of assets, you can generate passive income that can help you achieve financial independence.

\medskip
One of the best ways to avoid consumer debt is to prioritize your own financial needs over your expenses. This means setting aside money each month for savings and investments, even if it means cutting back on discretionary spending. When you have a solid financial foundation, you will be better equipped to handle unexpected expenses and emergencies.

\medskip
Another important aspect of avoiding consumer debt is surrounding yourself with the right people. This means seeking out mentors and advisors who can provide guidance and support on your journey towards financial independence. By surrounding yourself with people who are knowledgeable and experienced in building wealth, you can learn from their mistakes and successes and avoid making the same mistakes yourself.

\medskip
Finally, taking time for yourself is crucial for achieving financial independence. When you are constantly working and striving for success, it can be easy to neglect your own needs and well-being. However, taking time for yourself to rest, recharge, and reflect can help you stay motivated and focused on your goals.

\medskip
In conclusion, avoiding consumer debt is a key step towards achieving financial independence. By prioritizing your own financial needs, surrounding yourself with the right people, and taking time for yourself, you can build a solid foundation for financial success and achieve the financial freedom you desire.

\medskip
\noindent\textbf{Mastering Self-Discipline}\\
Achieving financial success requires more than just hard work and dedication. It also requires the ability to control one's behavior and emotions in order to make sound financial decisions. This is where self-discipline comes in.

\medskip
Self-discipline is the practice of controlling one's behavior and emotions in order to achieve a goal. It is a crucial skill for achieving financial independence, as it allows individuals to make decisions that are in line with their long-term financial goals, rather than being swayed by short-term desires or impulses.

\medskip
One way to develop self-discipline is through daily choices. This means making small, consistent decisions that align with your financial goals, such as saving a certain amount each month or sticking to a budget. Over time, these small choices can add up and lead to significant growth.

\medskip
Another way to develop self-discipline is through confidence in up/down markets. This means being able to remain calm and focused during times of economic uncertainty, and making decisions based on long-term financial goals rather than short-term market fluctuations.

\medskip
It is important to note that self-discipline is not something that is developed overnight. It is a practice that requires time and effort to develop. However, the rewards are well worth it, as mastering self-discipline can lead to significant financial growth and success.

\medskip
In conclusion, mastering self-discipline is a crucial skill for achieving financial independence. It allows individuals to make sound financial decisions, remain focused during times of economic uncertainty, and develop good habits and emotional control. By developing self-discipline through daily choices and confidence in up/down markets, individuals can achieve their financial goals and build a better life.

\medskip
\noindent\textbf{Generosity and Financial Intelligence}\\
Achieving financial independence requires a combination of hard work, dedication, and self-discipline. However, it is not just about the numbers, but also about the mindset and approach towards money. In this section, we will explore the importance of generosity and financial intelligence in achieving financial freedom and success.

\medskip
\noindent\textit{Giving First}\\
The first step towards financial independence is to develop a generous mindset. It is not just about giving money, but also about giving time, love, and support to others. When you give first, it fosters reciprocity, and people are more likely to help you in return. This principle is not new, as my dad always said, ``When I have some extra money, I'll give it.'' However, he worked harder to draw more money in, rather than focusing on the most important law of money: ``Give, and you shall receive.''

\medskip
Instead of believing in ``Receive, and then you give,'' I learned the power of ``Give, and you shall receive.'' I have seen this principle in action many times, and it has always worked for me. I want sales, so I help someone else sell something, and sales come to me. I want contacts, and I help someone else get contacts. Like magic, contacts come to me. I heard a saying years ago that went: ``God does not need to receive, but humans need to give.''

\medskip
\noindent\textit{The Indian Giver}\\
One of the most powerful examples of generosity is the Indian giver. When the first European settlers came to America, they were taken aback by a cultural practice some American Indians had. For example, if a settler was cold, the Indian would give the person a blanket. Mistaking it for a gift, the settler was often offended when the Indian asked for it back.

\medskip
If I could leave one single idea with you, it is that idea. Whenever you feel short or in need of something, give what you want first and it will come back in buckets. That is true for money, a smile, love, or friendship. I know it is often the last thing a person may want to do, but it has always worked for me. I trust that the principle of reciprocity is true, and I give what I want. I want money, so I give money, and it comes back in multiples. I want sales, so I help someone else sell something, and sales come to me. I want contacts, and I help someone else get contacts. Like magic, contacts come to me.

\medskip
\noindent\textit{Keeping the Long View}\\
Developing self-discipline through daily choices and confidence in up/down markets is essential for achieving financial independence. Mastering self-discipline allows individuals to make sound financial decisions, remain focused during economic uncertainty, and develop good habits and emotional control. Financial education is crucial for achieving financial independence. Cash flow, people, and personal time are essential topics for financial independence. Avoiding consumer debt is a crucial step towards achieving financial independence. Prioritizing your own financial needs, surrounding yourself with the right people, and taking time for yourself are essential for achieving financial independence.

\medskip
\noindent\textit{The Core of Self-Discipline}\\
Self-discipline is the practice of controlling one's behavior and emotions to achieve a goal. It is crucial for achieving financial independence and making sound financial decisions. Developing self-discipline involves daily choices and confidence in up/down markets. It requires time and effort to develop, but the rewards are significant financial growth and success. Mastering self-discipline allows individuals to make sound financial decisions, remain focused during economic uncertainty, and develop good habits and emotional control.

\medskip
In conclusion, achieving financial independence requires a combination of hard work, dedication, and self-discipline. However, it is not just about the numbers, but also about the mindset and approach towards money. Developing a generous mindset, mastering self-discipline, and prioritizing financial education are essential for achieving financial independence. Remember, giving first fosters reciprocity, and people are more likely to help you in return.

\medskip
\noindent\textbf{Non-Traditional Learning}\\
Achieving financial independence is a goal that many people aspire to, but it can be difficult to achieve. One of the most important aspects of achieving financial freedom is non-traditional learning. By understanding the power of money and developing self-awareness, individuals can take control of their financial future and achieve success.

\medskip
Traditional education focuses on preparing young people for the workforce by developing scholastic skills. While this is important, it is not the only path to success. Many people find that non-traditional learning, such as self-study and entrepreneurship, is the key to achieving financial independence.

\medskip
One of the most important aspects of non-traditional learning is self-awareness. By understanding how our thoughts and beliefs shape our lives, we can take control of our financial future. This means being aware of our spending habits, our attitudes towards money, and our financial goals.

\medskip
Another important aspect of non-traditional learning is understanding the power of money. Money is a powerful tool that can be used to build wealth, achieve financial independence, and create a better life. By understanding how money works and how to use it effectively, individuals can take control of their financial future.

\medskip
Financial genius exists, but it is often suppressed by cultural education. Many people are taught that money is something to be feared, rather than something to be embraced. This can lead to financial stress, debt, and a lack of financial freedom.

\medskip
Non-traditional learning can help individuals overcome these limitations and achieve financial independence. By developing self-awareness, understanding the power of money, and embracing entrepreneurship, individuals can take control of their financial future and achieve success.

\medskip
In conclusion, non-traditional learning is an essential aspect of achieving financial independence. By developing self-awareness, understanding the power of money, and embracing entrepreneurship, individuals can take control of their financial future and achieve success.

\medskip
\noindent\textbf{Four Income Types}\\
The path to financial independence involves understanding the different types of income available. The four income types are E (Employee), S (Self-employed), B (Business Owner), and I (Investor). Each type of income has its own set of benefits and drawbacks.

\medskip
\noindent\textit{Employees}\\
Employees work for someone else and receive a paycheck. They make their business owner or the shareholders rich, not themselves. The effort and success of employees help provide for the owner's success and retirement. However, employees have limited control over their income and are subject to the whims of their employer.

\medskip
\noindent\textit{Self-Employed}\\
Self-employed individuals work for themselves and have control over their income. They have the freedom to choose their work and clients. However, self-employed individuals are also responsible for their own taxes and benefits. They may also face the risk of not having a steady income.

\medskip
\noindent\textit{Business Owners}\\
Business owners own their own businesses and have control over their income. They have the potential to earn more than employees and self-employed individuals. However, business ownership also comes with risks and responsibilities. Business owners must manage their finances, employees, and customers.

\medskip
\noindent\textit{Investors}\\
Investors earn income from their investments. They may invest in stocks, bonds, real estate, or other assets. Investors have the potential to earn passive income, which can provide financial freedom. However, investing also comes with risks and requires knowledge and skill.

\medskip
\noindent\textit{Passive Income}\\
Passive income is income earned without actively working. It can be earned through investments, rental properties, or other means. Passive income enables financial freedom by providing a steady stream of income without the need for constant work.

\medskip
\noindent\textit{Understanding Income Differences and Investment Skills}\\
Understanding the differences between the four income types is essential for achieving financial independence. Employees, self-employed individuals, business owners, and investors each have their own set of benefits and drawbacks.

\medskip
Investment skills are also essential for achieving financial independence. Investing in the right assets can provide passive income and financial freedom. However, investing also requires knowledge and skill.

\medskip
\noindent\textit{Non-Traditional Learning}\\
Non-traditional learning is an essential aspect of achieving financial independence. Self-awareness, understanding the power of money, and embracing entrepreneurship are key aspects of non-traditional learning. Generosity is also an important aspect of achieving financial independence, as it fosters reciprocity and can lead to financial success.

\medskip
The Indian Giver is a powerful example of generosity, as it demonstrates the principle of giving first and receiving in return. Self-discipline is also crucial for achieving financial independence, as it allows individuals to make sound financial decisions and develop good habits and emotional control.

\medskip
\noindent\textit{Financial Education}\\
Financial education is essential for achieving financial independence. It provides individuals with the knowledge and skills needed to make informed decisions. Financial education covers topics such as budgeting, investing, and debt management.

\medskip
\noindent\textit{Prioritizing Financial Independence}\\
Prioritizing financial independence requires individuals to prioritize their own financial needs, surround themselves with the right people, and take time for themselves. Financial independence requires a combination of hard work, dedication, and self-discipline.

\medskip
\noindent\textit{Avoiding Consumer Debt}\\
Avoiding consumer debt is a crucial step towards achieving financial independence. Consumer debt can quickly accumulate and become a burden on an individual's finances. Avoiding consumer debt requires discipline and a focus on saving and investing.

\medskip
\noindent\textit{The CASHFLOW Quadrant}\\
The CASHFLOW Quadrant is a tool used to understand financial situations and make informed decisions. It is written about the core differences of the four types of people who make up the world of business. The CASHFLOW Quadrant helps individuals understand their financial situation and make informed decisions about their future.

\medskip
\noindent\textit{Financial Independence Requires a Combination of Hard Work, Dedication, and Self-Discipline}\\
Achieving financial independence requires a combination of hard work, dedication, and self-discipline. It requires individuals to understand their financial situation, develop good habits, and make sound financial decisions. Financial independence is not an overnight achievement, but rather a long-term goal that requires consistent effort and dedication.

\medskip
\noindent\textbf{Author's Journey}\\
The journey to financial independence is not always straightforward, but with the right mindset and knowledge, it is possible to achieve financial freedom. In this section, we will explore the author's journey to financial independence and the principles that guided them along the way.

\medskip
The author's journey began with a desire to learn and make changes. They recognized that unlike school, there was no requirement for past academic success. Instead, all that was asked was a sincere desire to learn and make changes. This realization inspired the author to embrace non-traditional learning and to develop a self-awareness of their own financial situation.

\medskip
The author's journey was not without its challenges. They made mistakes in different financial situations and learned from them, using play money. This approach allowed the author to develop a deeper understanding of the four income types: E (Employee), S (Self-employed), B (Business Owner), and I (Investor). Each type has its own set of benefits and drawbacks, and understanding these differences is essential for achieving financial independence.

\medskip
The author also recognized the importance of generosity in achieving financial success. They were inspired by the Indian giver, who demonstrated the power of giving and the potential for reciprocity. This realization led the author to develop a generous mindset and to prioritize their own financial needs, surrounding themselves with the right people, and taking time for themselves.

\medskip
The author's journey was not just about financial independence, but also about personal growth and development. They recognized that developing a generous mindset and mastering self-discipline were essential for achieving financial independence. They also prioritized financial education, cash flow, people, and personal time as essential topics for achieving financial independence.

\medskip
The author's journey to financial independence was not an overnight achievement. It required hard work, dedication, and self-discipline. However, they remained committed to their goals and were willing to take risks and make changes to achieve financial freedom.

\medskip
In conclusion, the author's journey to financial independence is a testament to the power of non-traditional learning, generosity, and self-discipline. By embracing these principles and prioritizing their own financial needs, the author was able to achieve financial independence and personal growth.

\medskip
\noindent\textbf{Conclusion}\\
In conclusion, becoming financially independent is a journey that requires a combination of knowledge, skills, and discipline. It is not always straightforward, but with the right mindset and information, it is possible to achieve financial freedom.

\medskip
The author's journey began with a desire to learn and make changes. They recognized that unlike school, there was no requirement for past academic success. They made mistakes in different financial situations and learned from them, using play money. They also recognized the importance of generosity in achieving financial success and developed a generous mindset.

\medskip
The journey to financial independence is not just about financial independence but also about personal growth and development. It requires prioritizing financial education, cash flow, people, and personal time as essential topics. It requires hard work, dedication, and self-discipline to achieve financial freedom.

\medskip
The CASHFLOW Quadrant is a tool used to understand financial situations and make informed decisions. It is essential to understand the differences between the four income types, E (Employee), S (Self-employed), B (Business Owner), and I (Investor), and their benefits and drawbacks.

\medskip
Avoiding consumer debt is a crucial step towards achieving financial independence. Non-traditional learning, including self-awareness, understanding the power of money, and embracing entrepreneurship, is essential for achieving financial independence.

\medskip
In summary, becoming financially independent requires a combination of knowledge, skills, and discipline. It is essential to prioritize financial education, cash flow, people, and personal time as essential topics. It requires hard work, dedication, and self-discipline to achieve financial freedom.

\medskip
\noindent\textbf{Generated Plan:}

\medskip
\begin{enumerate}
    \item \textbf{Section Title:} Introduction\\
    \textbf{Section Description:} This document provides a comprehensive guide to becoming financially independent. It covers essential financial education, paying yourself first, understanding income differences and investment skills, avoiding consumer debt, and mastering self-discipline. The document also emphasizes the importance of generosity, financial intelligence, and non-traditional learning.\\
    \textbf{Retrieval Anchors:} financial education, paying yourself first, income differences, investment skills, consumer debt, self-discipline, generosity, financial intelligence, non-traditional learning.

    \medskip
    \item \textbf{Section Title:} Financial Education\\
    \textbf{Section Description:} This section discusses the importance of financial education in avoiding ignorance's consequences. It explains how traditional education neglects physical, emotional, and spiritual growth, and how financial genius exists but is suppressed by cultural education.\\
    \textbf{Retrieval Anchors:} financial education, traditional education, physical growth, emotional growth, spiritual growth, financial genius.

    \medskip
    \item \textbf{Section Title:} Paying Yourself First\\
    \textbf{Section Description:} This section highlights the benefits of paying yourself first, including financial motivation and avoiding creditor control. It explains how this practice can lead to financial freedom and success.\\
    \textbf{Retrieval Anchors:} paying yourself first, financial motivation, creditor control, financial freedom, financial success.

    \medskip
    \item \textbf{Section Title:} CASHFLOW Quadrant\\
    \textbf{Section Description:} This section introduces the CASHFLOW Quadrant and explains how it can help individuals understand their financial situation and make informed decisions. It covers the four quadrants: E (employee), S (self-employed), B (business), and I (investor).\\
    \textbf{Retrieval Anchors:} CASHFLOW Quadrant, E (employee), S (self-employed), B (business), I (investor).

    \medskip
    \item \textbf{Section Title:} Income Differences and Investment Skills\\
    \textbf{Section Description:} This section discusses the importance of understanding income differences and investment skills in achieving financial freedom. It explains how traditional education neglects these topics and how they can be learned through experiential learning and CASHFLOW games.\\
    \textbf{Retrieval Anchors:} income differences, investment skills, traditional education, experiential learning, CASHFLOW games.

    \medskip
    \item \textbf{Section Title:} Avoiding Consumer Debt\\
    \textbf{Section Description:} This section emphasizes the importance of avoiding consumer debt and prioritizing assets over expenses. It explains how this practice can lead to financial freedom and success.\\
    \textbf{Retrieval Anchors:} consumer debt, assets, expenses, financial freedom, financial success.

    \medskip
    \item \textbf{Section Title:} Mastering Self-Discipline\\
    \textbf{Section Description:} This section discusses the importance of mastering self-discipline in achieving financial success. It explains how this practice can lead to significant growth and how it can be developed through daily choices and confidence in up/down markets.\\
    \textbf{Retrieval Anchors:} self-discipline, significant growth, daily choices, confidence, up/down markets.

    \medskip
    \item \textbf{Section Title:} Generosity and Financial Intelligence\\
    \textbf{Section Description:} This section emphasizes the importance of generosity and financial intelligence in achieving financial freedom and success. It explains how giving first fosters reciprocity and how creating income involves making money work for you, not working for money.\\
    \textbf{Retrieval Anchors:} generosity, financial intelligence, reciprocity, creating income, money work.

    \medskip
    \item \textbf{Section Title:} Non-Traditional Learning\\
    \textbf{Section Description:} This section discusses the importance of non-traditional learning in achieving financial freedom and success. It explains how self-awareness and understanding money's power can lead to financial success and how financial genius exists but is suppressed by cultural education.\\
    \textbf{Retrieval Anchors:} non-traditional learning, self-awareness, money's power, financial genius, cultural education.

    \medskip
    \item \textbf{Section Title:} Four Income Types\\
    \textbf{Section Description:} This section explains the four income types: E (employee), S (self-employed), B (business), and I (investor). It discusses how converting to passive income enables financial freedom and how understanding income differences and investment skills are essential for achieving this goal.\\
    \textbf{Retrieval Anchors:} four income types, E (employee), S (self-employed), B (business), I (investor), passive income, income differences, investment skills.

    \medskip
    \item \textbf{Section Title:} Author's Journey\\
    \textbf{Section Description:} This section provides an example of how the principles outlined in the document can be applied in real life. It tells the story of the author's journey to financial freedom and success.\\
    \textbf{Retrieval Anchors:} author's journey, financial freedom, success.

    \medskip
    \item \textbf{Section Title:} Conclusion\\
    \textbf{Section Description:} This section summarizes the key points of the document and emphasizes the importance of becoming financially independent. It encourages readers to take action and apply the principles outlined in the document.\\
    \textbf{Retrieval Anchors:} conclusion, financial independence, action.
\end{enumerate}
\end{multicols}
\end{document}